\documentclass[conference]{IEEEtran}
\IEEEoverridecommandlockouts
\usepackage{cite}
\usepackage{amsmath,amssymb,amsfonts}
\usepackage{algorithmic}
\usepackage{graphicx}
\usepackage{textcomp}
\usepackage{xcolor}
\def\BibTeX{{\rm B\kern-.05em{\sc i\kern-.025em b}\kern-.08em
    T\kern-.1667em\lower.7ex\hbox{E}\kern-.125emX}}
    
\makeatletter
\newcommand{\linebreakand}{%
  \end{@IEEEauthorhalign}
  \hfill\mbox{}\par
  \mbox{}\hfill\begin{@IEEEauthorhalign}
}
\makeatother

\begin{document}

\title{PRODUCT MARKET DEMAND ANALYSIS USING NLP IN BANGLISH TEXT WITH SENTIMENT ANALYSIS AND NAMED ENTITY RECOGNITION\\}

\author{\IEEEauthorblockN{Md Sabbir Hossain}
\IEEEauthorblockA{\textit{Department of Computer Science and Engineering} \\
\textit{Brac University}\\
Dhaka, Bangladesh \\
md.sabbir.hossain1@g.bracu.ac.bd}
\and
\IEEEauthorblockN{Nishat Nayla}
\IEEEauthorblockA{\textit{Department of Computer Science and Engineering} \\
\textit{Brac University}\\
Dhaka, Bangladesh \\
nishat.nayla@g.bracu.ac.bd}
\linebreakand
\IEEEauthorblockN{Annajiat Alim Rasel}
\textit{Lecturer}\\
\IEEEauthorblockA{\textit{Department of Computer Science and Engineering} \\
\textit{Brac University}\\
Dhaka, Bangladesh \\
annajiat@gmail.com}
}

\maketitle

\begin{abstract}
Product market demand analysis plays a significant role for originating business strategies due to its noticeable impact on the competitive business field. Furthermore, there are roughly 228 million native Bengali speakers, the majority of whom use Banglish text to interact with one another on social media. Consumers are buying and evaluating items on social media with Banglish text as social media emerges as an online marketplace for entrepreneurs. People use social media to find preferred smartphone brands and models by sharing their positive and bad experiences with them. For this reason, our goal is to gather Banglish text data and use sentiment analysis and named entity identification to assess Bangladeshi market demand for smartphones in order to determine the most popular smartphones by gender. We scraped product related data from social media with instant data scrapers and crawled data from Wikipedia and other sites for product information with python web scrapers. Using Python's Pandas and Seaborn libraries, the raw data is filtered using NLP methods. To train our datasets for named entity recognition, we utilized Spacey's custom NER model, Amazon Comprehend Custom NER. A tensorflow sequential model was deployed with parameter tweaking for sentiment analysis. Meanwhile, we used the Google Cloud Translation API to estimate the gender of the reviewers using the BanglaLinga library. In this article, we use natural language processing (NLP) approaches and several machine learning models to identify the most in-demand items and services in the Bangladeshi market. Our model has an accuracy of 87.99\%\ in Spacy Custom Named Entity recognition, 95.51\%\ in Amazon Comprehend Custom NER, and 87.02\%\ in the Sequential model for demand analysis. After Spacy's study, we were able to manage 80\%\ of mistakes related to misspelled words using a mix of Levenshtein distance and ratio algorithms.
\end{abstract}


\section{Introduction}
To thrive in a competitive market, the rapidly changing nature of today's economic climate necessitated a good business plan. If entrepreneurs want to start a new business in this highly competitive sector, they must first learn about the most popular gender-specific product on the market. It is critical to be familiar with the interests of the target customers to engage them with a sophisticated approach. Besides, sentiment analysis is a sufficient method to observe consumers preferences, desired models, and brands. Sentiment analysis effectively obtains, quantifies, reclaim and analyses consumers perceptions which benefits entrepreneurs to originate efficient business strategies.
Nowadays, social media has become one of the biggest online marketplaces for potential buyers and sellers. It allows entrepreneurs to engage and correlate with the interests of the consumers and study their behaviours. For the importance to economic and social development, sentiment analysis is now used in a variety of sectors, including business and social media marketing \cite{b1}. Our goal is to analyse the Banglish text data from social media buying-selling groups using natural language processing for a statistical and realistic market demand analysis for entrepreneurs based on smartphones. 
There is no such work with Banglish text for product market demand analysis. Hence, it was most challenging part for us to collect Banglish text data set. We collected raw Banglish text data from social platforms like buy and sell groups using data scraping tools. Another biggest challenge was to label the data set with its named entity and appropriate sentiment of demand based on Banglish text. For this reason, we applied natural language processing features to filter and cluster the data to turn it into more convenient format. Following, we applied machine learning algorithms (AWS Comprehend and Spacy NER) to train our data set and used the Sequential model from TensorFlow for performing sentiment analysis. Subsequently, validated accuracy of AWS Comprehend, Spacy NER and Sequential Model. These trained models and test data set provide a proper output of named entity and demand analysis. Further, with these model's output we also predicted gender from names. To sum up, we proposed a predictive model collecting raw data from social platforms, applying natural language processing and multiple data science algorithms to predict the market demand of smartphones based on consumer review for the entrepreneurs to have a crystal-clear knowledge about the competitive market.

\section{Related Work}
Market demand analysis plays a significant role in the economic environment as it helps to determine business policies. Text summarization can condense elaborated reviews into short sentences conveying the same idea and according to a research, combining the seq2seq model with the LSTM and attention mechanism can be effective for text summarizing. Their model uses multiple types of text summerization techniques based on input types, output types and the purposes \cite{b2}. 
Stock markets play a vital role in the economy but predicting and analysing it is not an easy task as a lot of factors are involved with it. Although, a machine learning algorithm can easily record and analyses the data’s keeping with all the significant factors with LSTM prediction on the MSE value, indicating that the LSTM model is efficient in time-series prediction, such as stock price and stock return \cite{b3}. As investors are familiar with behavioural finance of the consumers, sentiment analysis and natural language processing techniques can facilitate the selection of the most demand-able product in the market more efficiently by classifying positive and negative data \cite{b4}.
The effect of social platforms as a new emerging media on financial markets plays a significant role and sentiment revealed through social platforms has a larger and longer-lasting impact on market demand analysis \cite{b1}. The sentiment of the public in various social platforms as input of forecasting framework developed by eight regression models can be effective and among them fuzzy neural network based SOFNN algorithm give the higher accuracy of sentiment analysis \cite{b5}. Peoples emotions, attitudes, and opinions can be driven by applying multiple natural language processing features with sentiment analysis, these exacts the percentage of positive, negative, and neural from social media trending discussions. \cite{b6}. In e-commerce product reviews for the Bangla language, a sentiment detection system can be used for sentiment analysis, better decision-making, and improvements in products and services \cite{b7}. Large and microscopic companies are appearing in social networking sites to share their products and taking reviews from the customers and by using sentiment analysis they embrace the consumer interests and demands. A paper derives the accuracy of 85.25 \% performing sentiment analysis using natural language processing in Twitter data \cite{b8}.
Alongside, Bangla is a very complex language for embedding and clustering its words with wealthy literature developing on word embedding strategies and there is plenty of scope of improvement in Bangla language processing \cite{b9}. In a prior study KNN, Decision Tree, Random Forest performed well with similar accuracy but SVM and Logistic Regression gives higher accuracy for sentiment analysis in Bangla language \cite{b10}. Online engagement of Bangla language in the business sector is increasing day by day and it is very rational to use sentiment analysis of positive and negative feedback written in Bangla language performing natural language processing and five traditional machine learning algorithms to perform higher precisions \cite{b11}. Nowadays the usage of acronyms and abbreviations on social media is increasing and this is broadly used by teenagers and young adults. Banglish text is very flexible and a convenient way of communication for all kinds of people and online sellers and buyers are also communicating with this and using CNN and NLP features Banglish text sentiment can be acquired \cite{b12}. Online opinions are changing the way businesses are conducted, and a large amount of data is generated each year that is underutilized \cite{b13}. A machine learning based system can help customers have a better online shopping experience by allowing them to go through the system for product reviews based on the ratio of positive and negative feedback from previous customers \cite{b14}.
From the above discussion, it is clearly observed that most of the research in this field designed to predict stock price prediction and most of the Bangla language works are based on sentiment analysis. That is why we decided to work on product market demand analysis based on social media Banglish text using machine learning algorithms.

\section{Methodology}
This section represents our proposed work which focuses on a strategy to sentiment analysis on public Facebook groups, page’s comments data and their gender prediction for detecting most demanding device entities. The architectural overview describing an overall process design of sentiment analysis, gender prediction and named entity recognition which is shown in Fig. [1] . The developed method is based on several parts that are data scraping from social media, scrapping valid product name entities from authentic sites, pre-processing of the extracted social-media data using Natural Language tool-kits and Regular Expressions. To begin our research we need the clean device names list. To get that, firstly after scraping data from authentic sources we pre processed those using pandas, manual coding and regex-match and generated a phone list csv file so that we may use this in our research. 
models. 
\begin{figure}[h]
    \centering
    \includegraphics[width=0.5\textwidth]{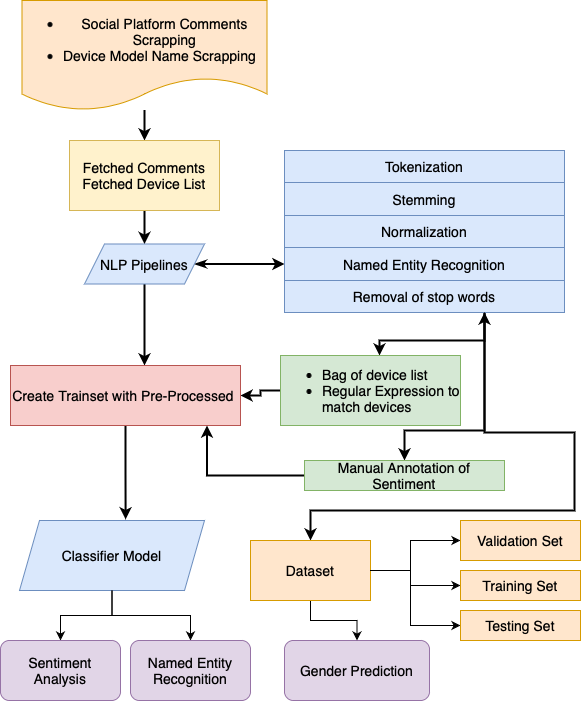}
    \caption{Work model flowchart}
    \label{fig : 4}
\end{figure}
To pre-processing and simplify the datasets firstly we have dropped all unnecessary columns and kept only the name and comments column. Secondly, we visualized the dataset with SeaBorn heatmap and dropped all null and duplicate values. We need a trainset to train sentiment analysis classifiers and named entity recognition classifiers. We splitted the datasets into train, test and validation sets ensuring each set contains at least 1000 valid data. Further, we labeled the sentiment manually for the training set so that we may get better results. After that, we labeled the name entity from comments. We defined a function to match the device names by comparing between the phone list and the comments data while ignoring the case and slight spelling errors with a combination of levenshtein ratio and edit distance algorithms. For predicting the gender, we have implemented a pre-trained model from BanglaLinga.As the predefined model doesn’t support Banglish names, Banglish text has been translated into Bangla names. Most of the pretrained models support gender detection from Bangla text. Although, there are not enough labeled Banglish datasets to predict gender from a Banglish name. Thus we found a feasible solution for our problem to translate Banglish to Bangla using the Google Cloud Translation api. We implemented the endpoint in our notebook successfully. Later, we feed the Bangla text into the pre-trained gender prediction model and it successfully gives the desired output. Though it has some exceptions like, it can only predict accurately with one word. So we had to send the prefix of the name. Consequently, we got too many exceptions like names starting with ‘Md’, ‘Phd’, ‘Dr’, ‘Mrs’, ‘Miss’, ‘Engr’ etc. We had to handle those exceptions using some cases with the help of regular expressions. Lastly, we have the clean train set and validation set. We made our sentiment analysis classifier with the Sequential model from keras. Then, we trained the model with the manual annotated train set of 3300 data with dropout value of 0.25. After fitting the model we moved to train our named entity classifier. Firstly we trained the NER model from Spacy. We fine tuned the parameters and got a satisfactory result. Secondly we used the Amazon Comprehend for custom named entity recognition. We trained the Comprehend with our labeled annotated sets and got more satisfactory results. Finally there are some misspelled device keywords which both models predict as incorrect. We made a combination of Levenshtein ratio and Edit Distance algorithms to correct the misspelled predicted devices name. Finally we plotted the most demanding device list based on the gender in the current market. 
\section{Dataset}
We made two data set for training our model. One is based on the customer product review and the other is device list data from Wikipedia. Banglish Text does not belong to any particular language, it is a decomposed form of Bangla language with English alphabets. Similarly, there are not that much work has been done with Banglish Texts. For this reason, we could not find organized, supervised data related to customer product review. Thus, we collected raw data and made a supervised data set using natural language processing features. Our data set contains more than 10000 raw data collected from social media. After pre-processing we have 5300 labeled data set with its sentiment, gender and product entity.
\subsection{Data Collection}
Nowadays, social networking sites play a phenomenal role in the producer-consumer relationship. There are a lot of mobile phone related groups or pages on social networking sites where consumers share their preferences, reviews. To get a proper analysis of product demand we decided to collect our raw data from social networking sites. Using the Instant Data Scraper tool we collected people’s Banglish comments related to mobile phones from various social networking sites. We stored the unsupervised data in a CSV file for further analysis. And for our second data set we collected smartphone model data from Wikipedia using python web scraper.
\begin{figure}[h]
    \centering
    \includegraphics[width=0.5\textwidth]{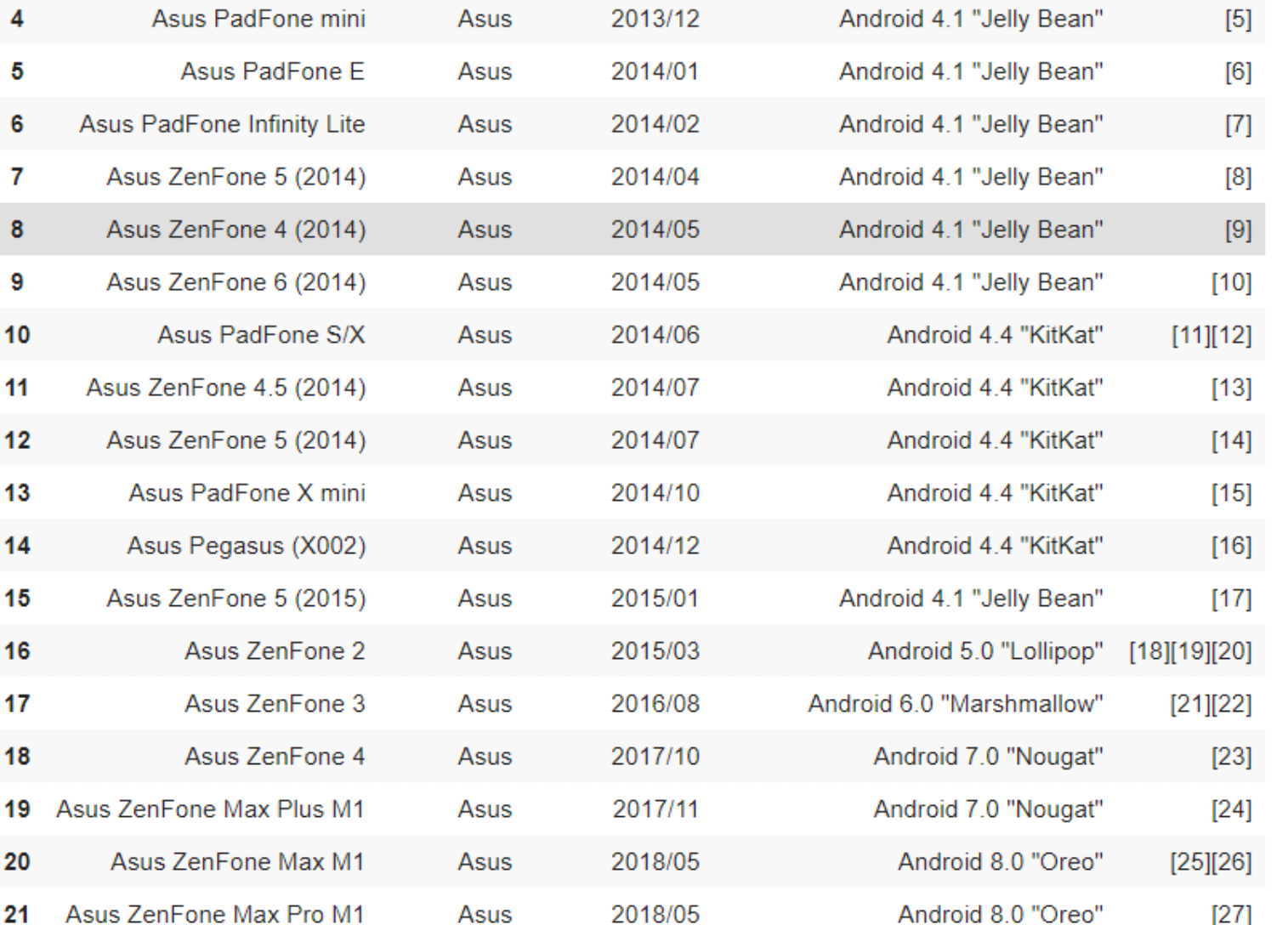}
    \caption{Collected dataset for the proposed model }
    \label{fig : 4}
\end{figure}
\subsection{Data Prepossessing}
We used an immediate data scraper to capture all of the information and saved it as several csv files. Multiple csv files were initially combined into a single large file. Then, based on column names, we examined the csv file and removed any superfluous columns. Another issue was duplicate data, which was eliminated using Python's Pandas Library. The Seaborn library was then used to check for null values. For sentiment analysis, we manually labeled the data set and utilized regular expressions with edit distance methods to identify entities.
Any device's name must be known in order for it to recognize device names from comments so we used pandas to compile a list of cellphones from Wikipedia. Then, we went through the data and chosen the columns which are required. We also included Apple devices list in the dataset from GSMArena. Finally, we created a CSV file with all smartphones names and brand.
The phone list data set have to be fine-tuned. The phone model includes the brand name as well as the model number, as can be seen. The device model "Galaxy S20" was, for example, "Samsung Galaxy S20" in the data set. People, on the other hand, do not leave comments or tweets containing the manufacturer's name. To solve this problem, we used a sophisticated approach depicted in the algorithm below to remove the developer's name from the device model.
We utilized regular expressions to eliminate certain indicators and the manufacturer's year from the data, as well as the developer's name. The device's model, on the other hand, had to be at least 7 characters long, including white spaces. If the developer name's character length falls below 7, we didn't cut it off. As a consequence, instead of "Apple iPhone XS ," we got "iPhone XS."
\begin{figure}[h]
    \centering
    \includegraphics[width=0.45\textwidth]{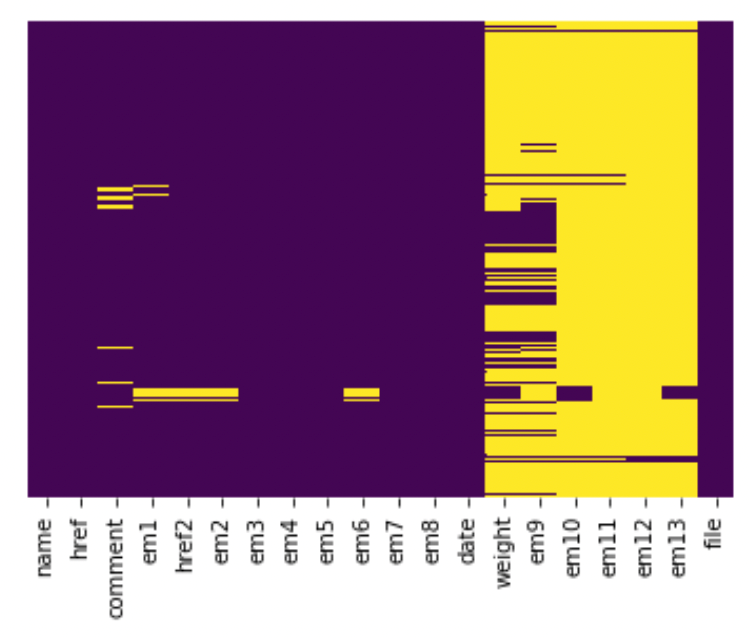}
    \caption{Data visualization explaining null values}
    \label{fig : 4}
\end{figure}

\subsection{Spell Correction Algorithm for Entities}
\begin{algorithmic}
\FORALL {$i$ in df.iterrows():}
    \STATE comment $\leftarrow$ str(df.comment[index])
    \STATE DeviceModels $\leftarrow$ wordTokenize(comment)
    \FORALL {$x$ in range(len(phone list)):}
       \STATE set MinDistance to $100$;
       \STATE set MaximumMatch to DeviceModels[x];
       \STATE set HighestRatio to $0$;
        \FORALL {$y$ in range(len(ModelName)):}
            \STATE d$\leftarrow$editdistance.eval(DeviceModels[x], str(ModelName[y]))
            \STATE 
            e=LevenshteinRatio(DeviceModels[x], str(ModelName[y]))
            \IF{d is less than 3 \\ and d is greater than equal to 0 \\ and d is less than MinDistance \\ and e is greater than 0.55 \\ and e is greater than HighestRatio } 
                \STATE set MinDistance to d
                \STATE set HighestRatio to e
                \STATE MaximumMatchv$\leftarrow$str(ModelName[y])
            \ENDIF
        \ENDFOR

        \STATE DeviceModels[x] $\leftarrow$ MaximumMatch
    \ENDFOR
    \STATE df.coment[index] $\leftarrow$ " ".join(DeviceModels)
\ENDFOR
\end{algorithmic}

\subsection{Train and Validation set}
We splitted our data into train set, test set and validation set. Our train set contains 3300 data. For refining the train set we dropped the rows which don't contain any device name. We normalized the dataset to its lower case. Further we used the stemming technique to noun entity using the edit distance algorithm and levenshtein ratio because there are not any predefined stemmers for Banglish corpora. Data labeling is performed using regular expression matches considering the case. Annotated sets were required for the Spacy custom NER and Amazon Comprehend model. We made two functions to make an annotated train set from train set so that we can feed this to Spacy custom NER and Amazon Comprehend custom named entity detection classifier. After that, we labeled each comment manually. Then we splitted the labeled dataset into train and test sets with 60:40 ratio. Finally our train and validation set is ready to fit and check the accuracy of our model. 
\section{Named Entity Recognition}
\subsection{Amazon Comprehend Custom NER Model}
Amazon Comprehend uses natural language processing (NLP) to extract insights about the content of documents. Amazon Comprehend processes any text file in UTF-8 format. Amazon Comprehend returns a list of entities, such as people, places, and locations, identified in a document. We made the annotated train set with the parameter of device name location in string.
We have also used AWS's built in custom named entity recognition model. We had train the model with lines of tweets and another CSV file containing the annotation for that lines of comments to train. We made a function to convert the trainset to both a text file which contains lines of comments and the CSV file which is the annotated documentation for the text file. We passed around 3380 labeled data set and trained the custom model within the web interface. We chose around 10\% of train data during the training phase to check the precision of the model. 
\subsection{SpaCy NER Model}
Spacy has the 'NER' pipeline component that identifies token spans fitting a predetermined set of named entities. For fitting our NER model we choose Spacy’s custom NER model with minibatch compound. For training purposes, it needs annotated labeled data. We made a function to make a trainable dataset for Spacy NER from a regular train set. We feed around 3300 data with its industrial level annotation. We chose the mini batch iteration of Spacy custom NER value of 5 and drop value of 0.1 and trained our custom model. It took several minutes to train in Google Colab.

\section{Gender Prediction}
Product demand varies from person to person. Preferences are mostly distinguished based on genders. It is very important to know the product demand based on gender. We implemented the gender prediction model from BanglaLinga. There were various errors like the model can’t properly predict gender based on full names. To handle this problem, we used the first name only. Even though various cases occurred with first names such as title like ‘Md’, ‘Phd’, ‘Dr’, ‘Mrs’, ‘Miss’, ‘Engr’ etc. We handled these exceptions with regular expression and forwarded the proper first name to the gender prediction model. The model even fails to predict gender from names from Banglish text. To solve the problem, we had to get the Google Cloud Platform and implement the cloud translation api. This cloud translation endpoint performed better than most other translation api in python. Finally we come up with the gender prediction part properly.
\section{Sentiment Analysis}
Sentiment analysis is a part of natural language processing which is also known as data mining. It extracts subjective information from a text and categorizes it into positive or negative. Sentiment Analysis (SA) is an opinion mining study that examines people's attitudes, sentiments, evaluations, and appraisals of societal entities such as businesses, persons, organizations and so on\cite{b15}.
A sequential model from tensorflow has been used for sentiment analysis. This model is best fitted for a plain stack of layers where each layer has exactly one tensor input and output. We set the pad sequences max length to 300, spatial dropout 1D to 0.25 and a dropout value for LSTM is 0.5. In our training we used the adam optimiser. For training the neural network model we chose the sigmoid activation function.

\begin{figure}[h]
    \centering
    \includegraphics[width=0.48\textwidth]{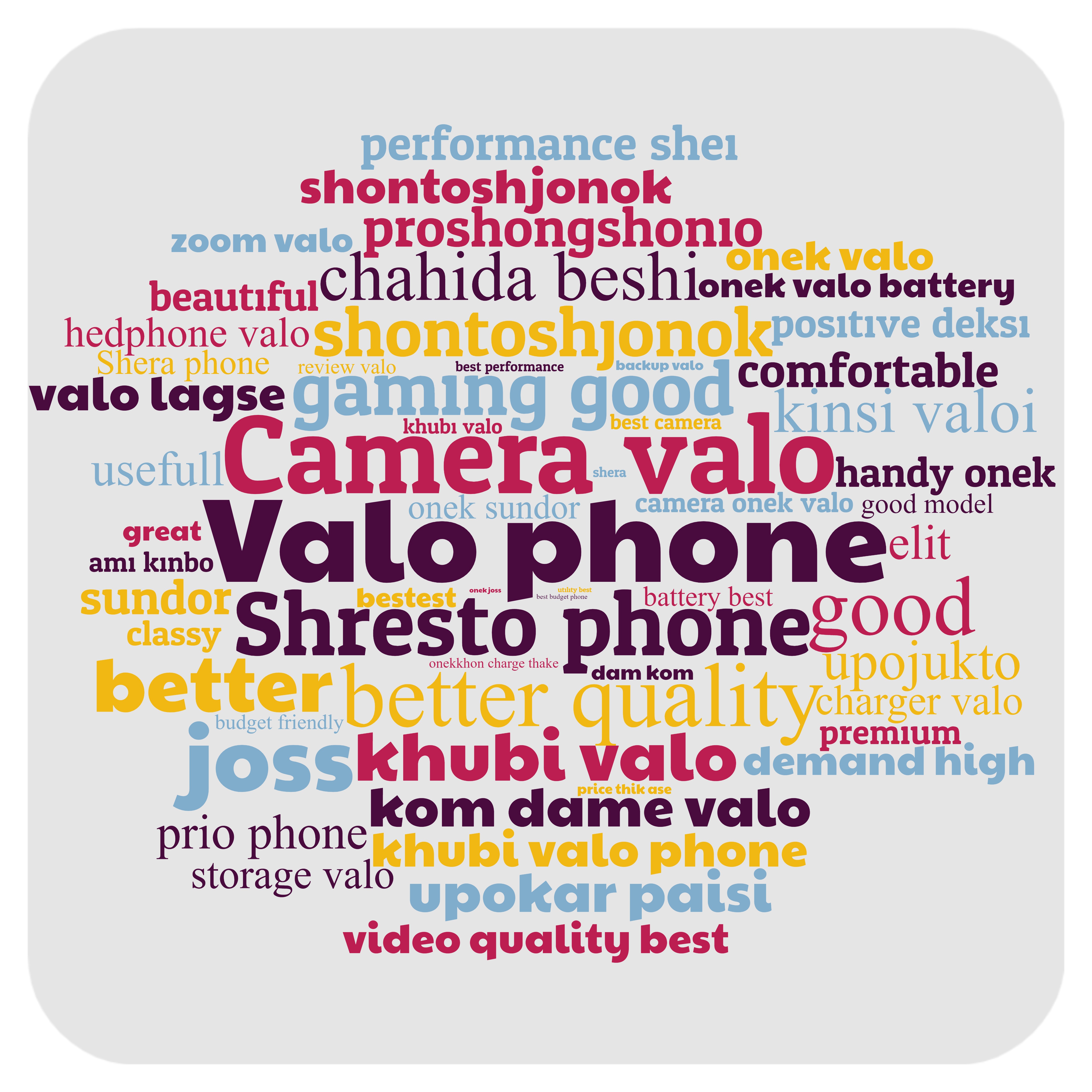}
    \caption{Positive Banglish text sentiment wordcloud }
    \label{fig : 4}
\end{figure}

\section{Result and Accuracy}
\subsection{Spacy Custom NER Accuracy}
We had a validation test set which consists of around 2000 labeled data. We matched every specific value with the Spacy NER result and got an outstanding accuracy of 87.99 percent. However we implemented some methods to fix spelling errors to increase the accuracy.
\subsection{Amazon Comprehend Accuracy}
Amazon Comprehend Custom Named Entity Recognition depicts an outstanding F1 score of 95.66 shown in Fig. [5].
To test the model we implemented the test set in similar fashion only without the annotation. It took around a few seconds to test 2000 data using our custom trained model and output the result as JSON format. We can see that it performs better than the Spacy custom NER system but it is not cost effective.

\begin{figure}[h]
    \centering
    \includegraphics[width=0.48\textwidth]{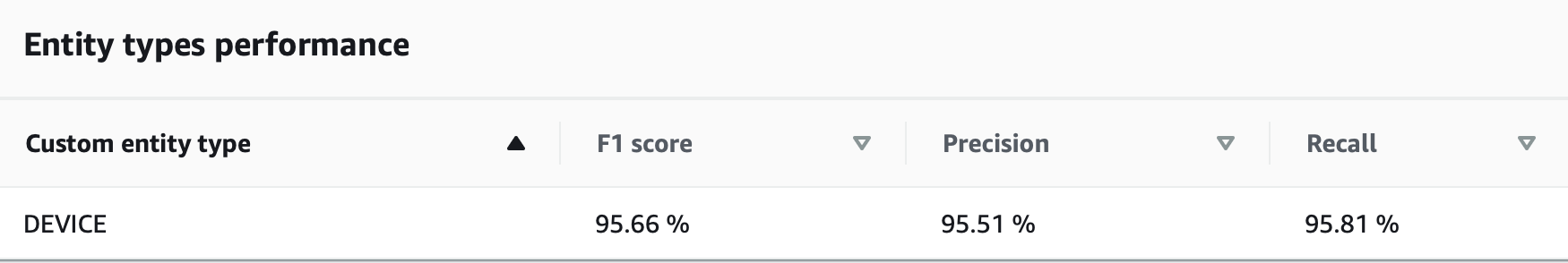}
    \caption{Amazon Comprehend Custom NER accuracy}
    \label{fig : 4}
\end{figure}
\subsection{Sequential Model Sentiment Analysis}
We tested our sentiment analysis model and found the peak floating number at where it differs from the positive or negative demand. We tuned the parameter for many times and tested with around 2000 datas and found the accuracy of 86.02 \% shown in Fig. [6].
\begin{figure}[h]
    \centering
    \includegraphics[width=0.35\textwidth]{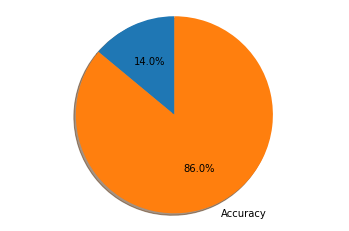}
    \caption{Sequential Model Accuracy}
    \label{fig : 4}
\end{figure}

\subsection{Demand Analysis}
For a specific period of time, market demand refers to how much people desire your goods. An rise in market demand occurs when more individuals seek a certain sort of goods. Plenty of stock is required under these conditions, and more individuals are ready to pay for it. We analyzed by the tweets positive and negative demand and tagged those with appropriate entity based on gender. Thus, we have sorted and plotted the demanding devices in descending manners which is one of the most useful tool for an entrepreneur. The Blue portion represents the ratio of male's demand where the orange portion represents the demand of device for a female shown in Fig. [7].
\begin{figure}[h]
    \centering
    \includegraphics[width=0.5\textwidth]{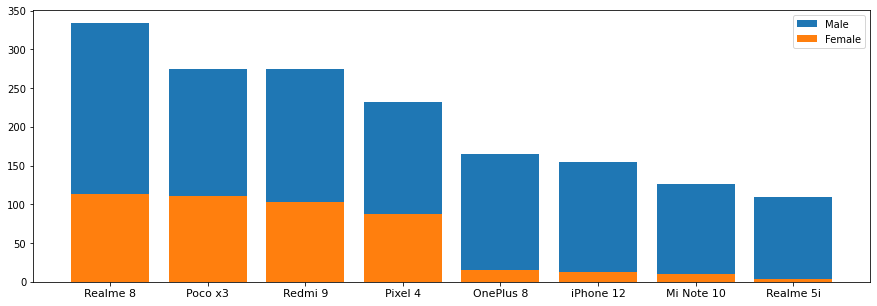}
    \caption{Demand analysis chart}
    \label{fig : 4}
\end{figure}
\section{Conclusion and future work}
In this paper, we used several machine learning algorithms such as the Spacy NER Model, Amazon Comprehend Custom Entity Recognition model. Then we used a Sequential model from keras for name entity recognition and we predicted gender from BanglaLinga library. Then we train and test our model with our customized data set. Through the name entity recognition model we successfully identified the gender of the person based on their names. Our model successfully identified the most demandable device model names from consumers' comments and posts data. Amazon's comprehensive model gives 95.66 \% accuracy on our data set. Then we performed sentiment analysis and it accurately predicted the positive reviewed devices. Finally, our project provides a graphical representation of most demanding and positively reviewed devices from current market scenarios based on gender.

\end{document}